%% file: main.tex
\documentclass[10pt,twocolumn,letterpaper]{article}

\usepackage{cvpr}              

\input{preamble}

\definecolor{cvprblue}{rgb}{0.21,0.49,0.74}
\usepackage[pagebackref,breaklinks,colorlinks,citecolor=cvprblue]{hyperref}
\usepackage{graphicx}
\usepackage{subcaption}
\usepackage{multirow}
\usepackage{comment}
\usepackage{booktabs}
\usepackage{amsmath}
\usepackage{amsfonts}
\usepackage{enumitem}
\usepackage{cuted}
\usepackage{makecell}

\makeatletter
\renewcommand{\maketag@@@}[1]{\hbox{\m@th\normalsize\normalfont#1}}%
\makeatother

\def\paperID{5831} 
\def\confName{CVPR}
\def\confYear{2024}

\title{Rethinking Evaluation Metrics of Open-Vocabulary Segmentaion}

\author{
Hao Zhou$^{1,3}$,
~~~
Tiancheng Shen$^{2}$,~~~
Xu Yang$^{3}$,~~~
Hai Huang$^{1}$,~~~ \\
Xiangtai Li$^4$,~~~ 
Lu Qi$^5$\thanks{Corresponding author.}, ~~~
Ming-Hsuan Yang$^5$, ~~~
\\[0.2cm]
$^1$Harbin Engineering University~~
$^2$The Chinese University of Hong Kong~~ \\
$^3$Institute of Automation, Chinese Academy of Sciences~~ \\
$^4$Nanyang Technological University~~
$^5$The University of California, Merced~~
}

\vspace{-0.4in}

\begin{document}
\maketitle
\input{sec/0_abstract}    
\input{sec/1_intro}
\input{sec/2_relatedwork}

\input{sec/3_metrics}
\input{sec/4_analysis}

\input{sec/5_benchmarks}

\input{sec/6_conclusion}
{
    \small
    \bibliographystyle{ieeenat_fullname}
    \bibliography{main}
}


\end{document}

%% file: preamble.tex
\usepackage[dvipsnames]{xcolor}


%% file: sec/0_abstract.tex
\vspace{-3mm}
\begin{strip}
  \centering
  \includegraphics[height=0.61\linewidth,width=\linewidth]{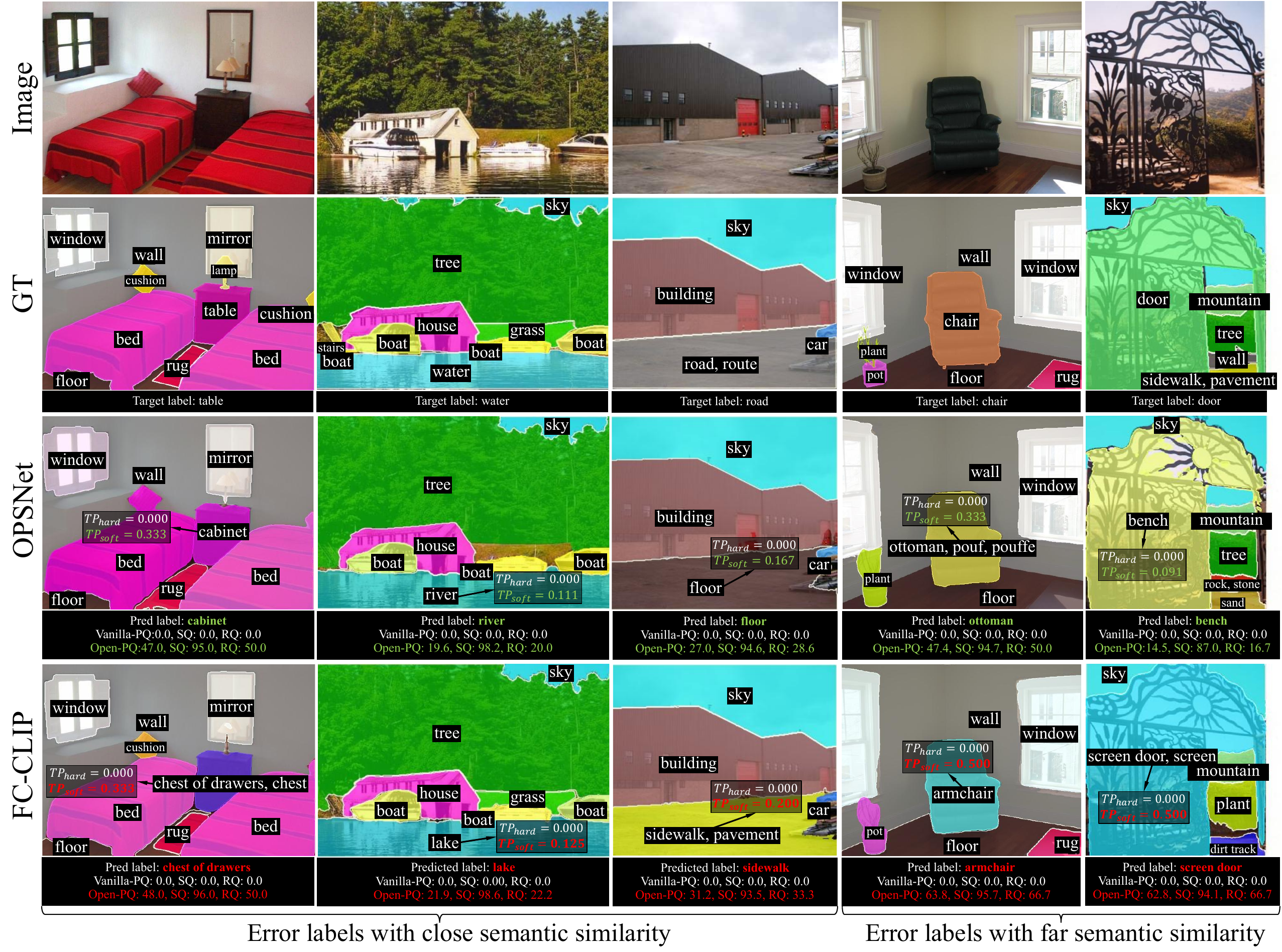}
  \vspace{-6mm}
  \captionof{figure}{Segmentation results with error labels evaluated by vanilla and open metrics. Open metric assigns soft true positive (TP) for masks with error labels and thus provides a more reasonable evaluation for open-vocabulary segmentation. 
}
\label{fig:teaser}
\end{strip}

\begin{abstract}
In this paper, we highlight a problem of evaluation metrics adopted in the open-vocabulary segmentation. That is, the evaluation process still heavily relies on closed-set metrics on zero-shot or cross-dataset pipelines without considering the similarity between predicted and ground truth categories.
To tackle this issue, we first survey eleven similarity measurements between two categorical words using WordNet linguistics statistics, text embedding, and language models by comprehensive quantitative analysis and user study. Built upon those explored measurements, we designed novel evaluation metrics, namely Open mIoU, Open AP, and Open PQ, tailored for three open-vocabulary segmentation tasks.
We benchmarked the proposed evaluation metrics on 12 open-vocabulary methods of three segmentation tasks. Even though the relative subjectivity of similarity distance, we demonstrate that our metrics can still well evaluate the open ability of the existing open-vocabulary segmentation methods. We hope that our work can bring with the community new thinking about how to evaluate the open ability of models. The evaluation code is released in \href{https://github.com/qqlu/Entity/tree/main}{github}.
\end{abstract}

%% file: sec/1_intro.tex
\section{Introduction}
\label{sec:intro}
Open-vocabulary image segmentation (OVS)~\cite{li2022language,ghiasi2021open,zhou2022maskclip} refers to the task of segmenting images into several regions while assigning each region a potentially \textit{unlimited or even open set} of object category. 
Compared to the traditional segmentation task that defines a fixed set of object classes, open-vocabulary segmentation is more flexible and adaptable to new and unseen object classes, which can benefit various risk-aware real-world applications like autonomous driving~\cite{qi2019amodal} and robot-related tasks~\cite{chu2021icm}. 

Recent developments in large vision language models (VLMs)~\cite{radford2021learning,li2022scaling}, such as CLIP~\cite{radford2021learning}, have significantly pushed the boundaries of the Open Vocabulary Segmentation (OVS) field~\cite{li2022language,ghiasi2021open,zhou2022maskclip}. However, the widely adopted evaluation methodologies, which often employ closed-set metrics on zero-shot or cross-dataset evaluations, fail to capture the open performance of OVS models without considering the similarity between predicted and ground truth categories. For example, 
the true positive (TP) score for both cabinet and chest would be zero if the ground truth is a table, as shown in the first column of Figure~\ref{fig:teaser}.
However, this contradicts human perceptual intuition~\cite{ungerleider1994and,bassett2011understanding} that we should not repudiate those predictions. This is our motivation why we should rethink how to estimate the model's open-world ability more reasonably.

To enable open-world evaluation, we first explore eleven heuristic similarity measurements between two categorical words that can be divided into three types, including WordNet methods, text-embedding models, and pre-trained language models. The WordNet is a lexical database that organizes words into a hierarchical structure by linguistician and is adopted by ImageNet~\cite{ILSVRC15}. Whereas the text embedding models (Word2Vec \cite{mikolov2013distributed}, GloVe \cite{pennington2014glove}, and FastText \cite{bojanowski2017enriching}) and pre-trained language models (CLIP \cite{radford2021learning}, BERT \cite{devlin2018bert}, GPT \cite{radford2019language}, and T5 \cite{raffel2020exploring}) are learnable models trained in language datasets at word and sentence level respectively. After the deep quantitative analysis and user study in Section~\ref{sec:analysis}, we prefer to use the path similarity of WordNet due to four advantages: 1) generates moderate similarity values. 2) able to distinguish polysemy. 3) unbiased to training data. 4) more aligned with human cognition. It is noted that this choice is preferred and still subjective, which might not be adaptable for each open-world situation.
%
%
%


Built upon the similarity measurements, we introduce Open mIoU, Open AP, and Open PQ to evaluate open-vocabulary semantic, instance, and panoptic segmentation tasks, respectively. Although these three segmentation tasks have distinct objectives and evaluation pipelines, our open metrics share two fundamental operations: a class-agnostic matching criterion and a semantic similarity-based scoring computation.
%
Initially, we first employ a class-agnostic matching approach, identifying the highest Intersection over Union (IoU) between the predicted regions and the ground truth. This step differs from the one in traditional evaluation metrics, where the matching process is limited to predicted and ground truth regions within the same category, even when their masks are highly overlapped. Subsequently, we utilize the similarity measurement explored above to compute floating-point matching scores instead of the binary true positive (TP), false positive (FP), and false negative (FN) values employed in close-set segmentation metrics.

Finally, we use our proposed evaluation metrics to benchmark 12 open-vocabulary segmentation methods of three segmentation tasks. By qualitatively comparing these new metrics to vanilla close-set metrics in Figure~\ref{fig:teaser}, we demonstrate that our novel metrics align more closely with human judgment, providing a more accurate and reasonable assessment of the model's open capability. The user study is further conducted to validate our argument.

We conclude our contributions as follows:
\begin{itemize}
	
\item We are the first to highlight a significant limitation of existing evaluation metrics applied to open-world segmentation. The predicted classes should be assigned soft match scores to the same ground truth based on their similarity distance.

\item We compare eleven similarity measurements using WordNet linguistics statistics, text embedding, or language models in comprehensive quantitative analysis and user studies.
	
\item We propose innovative open evaluation metrics, including Open mIoU, AP, and PQ. Then, the three mainstream open-vocabulary segmentation (semantic, instance, and panoptic segmentation) and object detection tasks and models are benchmarked. 

\item Even though the open evaluation metrics are still subjective, they can inspire the community to rethink about a reasonable way to judge the model's open ability.

\end{itemize}


%% file: sec/2_relatedwork.tex
\section{Related Work}
\label{sec:relatedwork}

\noindent
\textbf{Closed-Set Segmentation.} This task aims to segment images into regions with predefined class categories. The Fully Convolutional Networks (FCN)~\cite{long2015fully} opens the deep learning era for image segmentation.
Later, several convolution-\cite{li2023transformer} or transformer-based~\cite{liu2021swin} works further improve the model's ability in semantic segmentation~\cite{zhao2017pyramid,wang2020deep,shen2022high,chen2017deeplab,ronneberger2015u,xie2021segformer}, instance segmentation~\cite{he2017mask,lin2014microsoft,chen2019hybrid,liu2018path,qi2021pointins}, and panoptic segmentation~\cite{kirillov2019panoptic,kirillov2019panopticfpn,cheng2021mask2former,zhang2021knet}.
Although closed-set segmentation approaches continue to advance, the predefined category set is ill-posed for addressing the demands of real-world vision applications, such as scene understanding, self-driving, and robotics, where the number of object categories is extensive and subject to change. In our work, we focus on enabling the closed-set evaluation metrics to have open-world ability.

\noindent 
\textbf{Open-Vocabulary Segmentation.} At the earlier stage,~\cite{zhao2017open} proposes an open-vocabulary parsing network, which encodes the word concept hierarchy based on dictionaries and learns a joint pixel and word feature space. Based on the multi-modality model CLIP~\cite{radford2021learning} or ALIGN~\cite{jia2021scaling} trained on large-scale image-text pairs and class-agnostic entity proposals~\cite{cheng2021mask2former,qi2021open,qi2023high}. LSeg~\cite{li2022language}, MaskCLIP~\cite{zhou2022maskclip}, OpenSeg~\cite{ghiasi2021open}, OVSeg~\cite{liang2022open} and follows~\cite{ren2023prompt,chen2023open,zhang2023simple,wu2023betrayed,yu2023convolutions,xu2023open}  utilizes text embeddings from CLIP to associate the class-agnostic entity embeddings.

Even though these methods consider cross-modal feature representation and alignment, their evaluation still relies on closed-set metrics on zero-shot or cross-dataset settings. Thus, the existing open-vocabulary metrics still punish the inconsistency between predicted and ground truth categories even though they are synonyms. Instead, our work aims to design a more reasonable evaluation pipeline considering the similarity between two labels.

\noindent
\textbf{Open-Vocabulary Object Detection (OVOD).} Considering the close relationship between instance segmentation and object detection, several works~\cite{pb-ovd, xpm, ovr-cnn, detic, DetPro,Wang2023ObjectAwareDP, Minderer2022SimpleOO,Chen2022OpenVO,Wang2022LearningTD,Du2022LearningTP,gu2021open_vild, wu2023baron, DetPro,OV-DETR,Kuo2022FVLMOO, wu2023clipself, xu2023dst,kim2023contrastive,kim2023region} explore object detection in open-vocabulary settings, extending detectors to recognize objects not encountered during the training. 
Similar to previous segmentation tasks, the existing OVOD approaches still use closed-set metrics (box mAP). Our proposed open metric can also be adapted to OVOD settings.

%% file: sec/3_metrics.tex
\section{Evaluation Metrics}
\label{sec:metrics}

In this section, we initially present an overview of three prevalent closed-set evaluation metrics, namely, mean Intersection over Union (mIoU), Average Precision (AP), and Panoptic Quality (PQ). These metrics are widely used in semantic, instance, and panoptic segmentation tasks, with the AP metric also capable of evaluating object detection by replacing mask IoU with box IoU. Following this, we introduce the evaluation metrics we've proposed for open-vocabulary tasks within these domains.

\subsection{Preliminaries}
Despite the various segmentation tasks targeting different levels, such as pixel, object, and entity, we observe that the core of their evaluation metrics is the calculation of IoU at these levels, which measures the precision between the predicted results and ground truth. Usually, we can obtain a confusion matrix by dividing the predicted results into four sets, including True Positives (TP), True Negatives (TN), False Positives (FN), and False Negatives (FN), based on the matching criteria to the ground truth at each category. Specifically, TP and TN represent the count of accurate positive and negative predictions, respectively. Conversely, False FP and FN quantify incorrect positive and negative predictions, respectively. With these definitions in place, Intersection over Union (IoU) can be mathematically formulated by
${\rm IoU} = \frac{|TP_p|}{|FP_p|+|TP_p|+|FN_p|}$, where $|TP_p|$, $|FP_p|$, and $|FP_p|$ respectively stand for the number of pixel-level TP, FP, and FN. After that, we calculate the corresponding evaluation metrics given the different segmentation tasks.

For the mIoU metric used in semantic segmentation, the matching criterion is determined by the accurate pixel-level prediction corresponding to the ground truth classes. The mIoU is calculated as:
\begin{equation}
    {\rm mIoU} = \sum\nolimits_{i=0}^k{\rm IoU}_i,
\end{equation}
where ${\rm IoU}_i$ represents the IoU of class $i$ and $k+1$ is the total number of classes in the evaluated dataset. The AP and PQ metrics used in instance and panoptic segmentation rely on entity-level matches determined by IoU. These matches are evaluated based on whether the IoU between the predicted and ground truth masks surpasses a specified threshold $\delta$. Subsequently, the AP computes the precision-recall curve using entity-level TP, FP, and FN, which are defined as follows:
\begin{equation}
	\label{eq.2}
	{\rm AP} = \int_0^1 p(r)\mathrm{d}r,
\end{equation}
%
where $p=\frac{|TP_{e}|}{|TP_{e}|+|FP_{e}|}$ and $r=\frac{|TP_e|}{|TP_e|+|FN_e|}$ denotes precision and recall, while $|TP_{e}|$, $|FP_{e}|$ and $|FN_{e}|$ represent the numbers of entity-level TP, FP, and FN respectively. Meanwhile, the PQ evaluates the segmentation quality of the model as follows:
\par
\vspace{-3.5mm}
\begin{footnotesize}
\begin{equation}
    \hspace{-2.5mm}
	\label{eq.3}
	{\rm PQ} \!=\! \underbrace{\frac{\sum_{(g_i,d_j)\in TP}{\rm IoU}(g_i, d_j)}{|TP_{e}|}}_{\rm SQ} \!\times\! \underbrace{\frac{|TP_{e}|}{|TP_{e}|\!+\!\frac{1}{2}|FP_{e}|\!+\!\frac{1}{2}|FN_{e}|}}_{\rm RQ},
\end{equation}
\end{footnotesize}%
where ${\rm IoU}(d_i,g_j)$ represents the IoU between entity $d_i$ and $g_j$. The PQ can be interpreted as the product of the Segmentation Quality (SQ) and Recognition Quality (RQ).

\begin{table*}[tbp]
	\centering
        \resizebox{0.95\textwidth}{!}{
		\begin{tabular}{c|c|c|c|c}
			\toprule
			Metric & Type & Situation & Hard TP/FP/FN (vanilla) & Soft TP/FP/FN (open)\\
			\midrule
			mIoU & pixel-based & \makecell[c]{$g_i(x, y)=d_j(x, y)$ and $c_i \neq c_j$} & \makecell[c]{$TP_{c_i}=0,$\\$FP_{c_j}=1, FN_{c_i}=1$} & \makecell[c]{$TP_{c_i}=S_{{c_i}{c_j}}, FP_{c_j}=1-S_{{c_i}{c_j}},$\\ $FN_{c_i}=1-S_{{c_i}{c_j}}$}\\
			\midrule
			AP & entity-based & \makecell[c]{${\rm IoU}(g_i, d_j) \geq \delta$ and $c_i \neq c_j$} & \makecell[c]{$TP_{c_i}=0, FP_{c_j}=1$} & \makecell[c]{$TP_{c_i}=S_{{c_i}{c_j}}, FP_{c_j}=1-S_{{c_i}{c_j}}$}\\
			\midrule
		  PQ & entity-based & \makecell[c]{${\rm IoU}(g_i, d_j) \geq \delta$ and $c_i \neq c_j$ and \\ $(c_i, c_j\in stuff$ or $c_i, c_j \in thing)$} & \makecell[c]{$TP_{c_i}=0,$\\$FP_{c_j}=1, FN_{c_i}=1$} & \makecell[c]{$TP_{c_i}=S_{{c_i}{c_j}}, FP_{c_j}=1-S_{{c_i}{c_j}},$\\ $FN_{c_i}=1-S_{{c_i}{c_j}}, {\rm IoU}(g_i, d_j)*=S_{{c_i}{c_j}}$}\\
			\bottomrule
	\end{tabular}}
    \vspace{-2mm}
    \caption{Hard and Soft TP/FP/FN that respectively used in vanilla metrics and open metrics defined with the unified notation from Table \ref{tab:notation}.}
    \label{tab:definition}
	\vspace{-4mm}
\end{table*}

\begin{table}[tbp]
	\centering
        \resizebox{0.95\linewidth}{!}{
		\begin{tabular}{c|c}
			\toprule
			Notation & Definition \\
			\midrule
			$g_i$ & ground truth mask or pixel \\
            $d_j$ & prediction mask or pixel \\
			\midrule
			$g_i(x, y)$ & ground truth pixel's coordinates \\
   		$d_j(x, y)$ & prediction pixel's coordinates \\
			\midrule
		  $c_i$ & label of ground truth $g_i$\\
            $c_j$ & label of prediction $d_j$\\
            \midrule
            ${\rm IoU}(g_i, d_j)$ & IoU between ground truth and prediction masks \\
			\bottomrule
            $\delta$ & matching threshold for mask IoU \\
			\bottomrule
	\end{tabular}}
    \vspace{-2mm}
    \caption{Notation used in Table \ref{tab:definition}. We use $g_i$ and $d_j$ to simultaneously
 represent ground truth and prediction of masks or pixels for simplicity.}
    \label{tab:notation}
	\vspace{-4mm}
\end{table}

\subsection{Open-Vocabulary Evaluation}

The definitions of mIoU, AP, and PQ reveal that they are ill-equipped for assessing open-vocabulary segmentation tasks as they require strict class matching. To mitigate this limitation, we introduce open metrics that incorporate considerations of semantic similarity between labels. Enabling open evaluations involves tackling two crucial aspects. First, we must establish a reliable method for assessing the semantic similarity between ground truth and predicted labels. Then, based on the defined similarities, we must calculate soft versions of TP, FP, and FN to derive open metrics suitable for open-vocabulary segmentation tasks. 

\subsubsection{Similarity Matrix}

There are two prominent categories of models used extensively to quantify semantic similarity between words: distributional models and knowledge-based models. Knowledge-based models leverage structured knowledge resources, with WordNet~\cite{miller1995wordnet} being the most prevalent among them. Distributional models extract word embeddings from text embedding or pre-trained language models and subsequently compute their cosine similarity. Recognizing the enhanced capacity of pre-trained language models to comprehend sentences or documents, we further classify distributional models into two distinct groups. Consequently, there are three primary categories of methods: (1) WordNet methods, (2) Text embedding models, and (3) Pre-trained language models.

\noindent
\textbf{WordNet Methods.} WordNet is a lexical database that organizes words into a hierarchical structure using an "IS-A" (hypernym/hyponym) taxonomy, echoing how the human cognitive system stores visual knowledge, proves to be a valuable resource for assessing the semantic relationship between different categories. Therefore, plenty of methods~\cite{wu1994verb,lin1998information,lin1998information,lesk1986automatic,leacock1998combining,resnik1995using} are proposed to measure similarity based on WordNet. Among these methods, the Path~\cite{lesk1986automatic}, Wu-Palmer~\cite{wu1994verb}, Lin~\cite{lin1998information}, and Lesk~\cite{lesk1986automatic} Similarities yield similarity values within the closed interval $\left[0, 1\right]$, which is conducive for calculating soft TP (FP, FN), making them suitable choices.

\noindent
\textbf{Text Embedding Models.} Text embedding models which trained on massive text corpora learn to represent words in vector spaces where similar words are closer in the vector space, indicating their semantic similarity. Therefore, models like Word2Vec \cite{mikolov2013distributed}, GloVe \cite{pennington2014glove}, and FastText \cite{bojanowski2017enriching} are preferred for calculating word similarities using cosine similarity. Note that cosine similarity scores range from -1 to 1, where the interval [-1, 0) signifies negative correlation is redundant for calculating soft TP (FP, FN) scores. Hence, we set cosine similarity values less than 0 to 0.

\noindent
\textbf{Pre-trained Language Models.} Expect for the word-level embeddings ability, pre-trained language models also can comprehend the entire sentences or documents that text embedding models cannot, and thus can extract better embeddings for compound words. Hence, models such as CLIP \cite{radford2021learning}, BERT \cite{devlin2018bert}, GPT \cite{radford2019language}, and T5 \cite{raffel2020exploring} are also well-suited for calculating word similarities.


\noindent
\textbf{Solutions to Unknown Words.} Pre-trained language models are designed to understand the context of each word within a sentence, enabling them to generate embeddings for unknown words using relevant known words or sub-words. Text embedding models are trained to provide embeddings for individual words, not for phrases or sentences. When confronted with unknown words, we follow a standard procedure of initializing these words with random vectors. WordNet is one of the most extensive lexical dataset (contains 82,115 noun synsets) and undergoes continuous updates. For words absent in it, we choose the hypernym of these words as a replacement because hypernyms encompass most of the characteristics of their hyponyms, rendering minor changes in path length (either plus or minus one) of minimal consequence to the final evaluation results.


\noindent
\textbf{Our Preference.} In Section~\ref{sec:analysis}, we conduct a comprehensive analysis of the eleven similarity measurements mentioned above and prefer using Path Similarity to calculate the similarity matrix $S \in \mathcal{R}^{(k+1) \times (k+1)}$. For further information, please refer to Section~\ref{sec:analysis}.

\subsubsection{Open Evaluation Metrics}
Instead of assigning binary TP, FP, and FN values to predictions, we propose open evaluation metrics that convert these binary values into floating-point scores. There are two primary steps to achieve this target. The first step is the transition from a class-specific matching strategy to a class-agnostic one. The second step utilizes the Path Similarity to calculate soft TP, FP, and FN based on the matching conditions between the ground truth and predicted masks. In the following, we introduce the detailed evaluation process for each segmentation task.

\noindent
\textbf{Open mIoU.} Semantic segmentation is a pixel-level classification task primarily focused on category prediction at the pixel level. Traditional TP${_\text{hard}}$, FP${_\text{hard}}$, and TN${_\text{hard}}$ for class $i$ are calculated as $TP{_\text{hard}} = n_{ii}$, $FP_{\text{hard}} = \sum_{j=0}^k n_{ji} - n_{ii}$, and $FN_{\text{hard}} = \sum_{j=0}^k n_{ij} - n_{ii}$, where $n_{ij}$ symbolizes pixels with the ground truth label $i$ classified as label $j$. However, in open-vocabulary evaluation, misclassified pixels shouldn't be strictly counted as binary TP, FP, or FN. Instead, we suggest the calculation of soft TP, FP, and FN for error label prediction situations by utilizing the semantic similarity matrix $S$, as shown In table \ref{tab:definition}. Therefore, the soft TP, FP, and FN for class $i$ can be defined as below:
\par
\vspace{-3.5mm}
\begin{footnotesize}
\begin{equation}
    \hspace{-2mm}
    \begin{split}
		TP_{\text{soft}} \!&=\! c_{ii} \!+\! \sum_{j=0}^k S_{ij}\cdot c_{ij} \!-\! S_{ii}\cdot c_{ii} \!=\! \sum_{j=0}^{k+1} S_{ij}\cdot c_{ij}, \\
		FP_{\text{soft}} \!&=\! \sum_{j=0}^k (1 \!-\! S_{ji}) \cdot c_{ji} \!-\! (1 \!-\! S_{ii}) \cdot c_{ii} = \sum_{i=0}^{k+1} (1\!-\!S_{ji}) \cdot c_{ji}, \\
		FN_{\text{soft}} \!&=\! \sum_{j=0}^k (1 \!-\! S_{ij})\cdot c_{ij} \!-\! (1 \!-\! S_{ii})\cdot c_{ii} = \sum_{j=0}^{k+1} (1 \!-\! S_{ij})\cdot c_{ij},
	\label{eq.5}
    \end{split}
\end{equation}
\end{footnotesize}%
where $S_{ij}$ represents the similarity score between labels $i$ and $j$ in similarity matrix $S$. The notations $*_{\text{soft}}$ and $*_{\text{hard}}$ represent the soft and binary versions of $*$, where $*$ can be TP, FP, or FN. Afterward, we can calculate the Open mIoU for open-vocabulary semantic segmentation using the IoU formulation. 

\noindent
\textbf{Open AP.} In the context of instance segmentation, matching is performed between predicted masks and ground truth masks that share the same category label. To modify AP for open-vocabulary evaluation, we first convert the class-aware matching strategy to a class-agnostic manner. Specifically, we convert the labels of all the ground truth and predicted masks to the "object" category. Then, we sort the predicted masks in descending order of their prediction scores, creating a class-agnostic global ranking. Subsequently, we match the predicted masks using this global descending order. For each predicted mask, we match it with a ground truth mask that has the largest IoU and has not been assigned to any predictions. Let $g_i$ be a ground truth mask and $d_j$ be a predicted mask that satisfies the matching criteria but with different labels. According to the labels of $g_i$ and $d_j$, we calculate soft TP and FP using the similarity matrix $S$, as illustrated in Table \ref{tab:definition}. Given the soft TP and FP, the Open AP can be obtained following Equation \ref{eq.2}.

\noindent
\textbf{Open PQ.} The vanilla PQ metric performs matching between predicted and ground truth masks based on two criteria: the IoU threshold exceeding 0.5 and the requirement of belonging to the same category. To enable open-vocabulary evaluation, we only need to modify the category-matching criteria of PQ. Instead of strictly requiring two overlapped masks $g_i$ and $d_j$ to have the same category label, the labels of two masks only need to belong to stuff or thing classes. Then, we calculate the soft TP, FP, and FN using the semantic similarity between the labels of $g_i$ and $d_j$, as illustrated in Table \ref{tab:definition}. Finally, the Open PQ value can be obtained using Equation \ref{eq.3}. Note that the SQ term in PQ calculates the average IoU over matched segments. However, the match criteria for open-vocabulary segmentation have been updated to a softer manner that considers the semantic similarity between labels. Therefore, the ${\rm IoU}(g_i,d_j)$ in Equation \ref{eq.3} should be multiplied by the same semantic similarity to keep the definition of SQ unchanged. 

%% file: sec/4_analysis.tex
\section{Analysis of Different Similarity Measures}
\label{sec:analysis}

\begin{figure*}[!t] 
	\centering
	\begin{minipage}[b]{0.95\linewidth}
		\subfloat[All]{
			\begin{minipage}[b]{0.24\linewidth} 
				\centering
				\includegraphics[width=\linewidth,trim=20 20 25 25,clip]{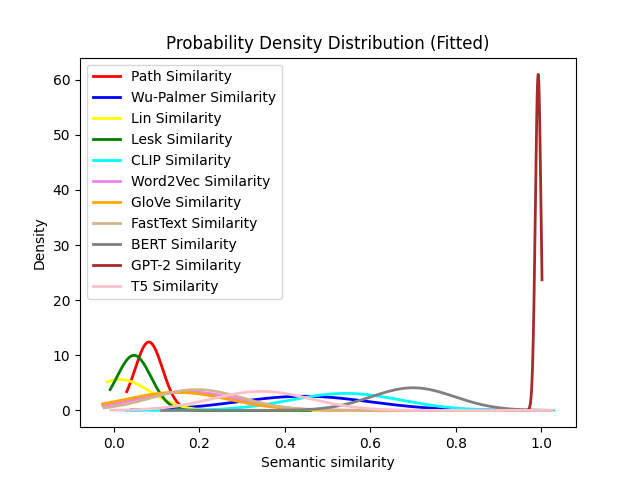}
			\end{minipage}
            \label{fig:pdf_all}
		}\hspace{-10pt}
		\hfill
		\subfloat[WordNet]{
			\begin{minipage}[b]{0.24\linewidth}
				\centering
				\includegraphics[width=\linewidth,trim=20 20 25 25,clip]{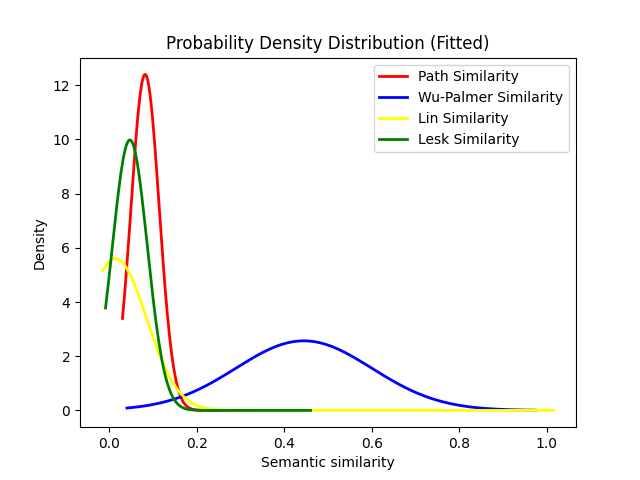}
			\end{minipage}
            \label{fig:pdf_wn}
		}\hspace{-10pt}
		\hfill
		\subfloat[Text embedding model]{
			\begin{minipage}[b]{0.24\linewidth}
				\centering
				\includegraphics[width=\linewidth,trim=20 20 25 25,clip]{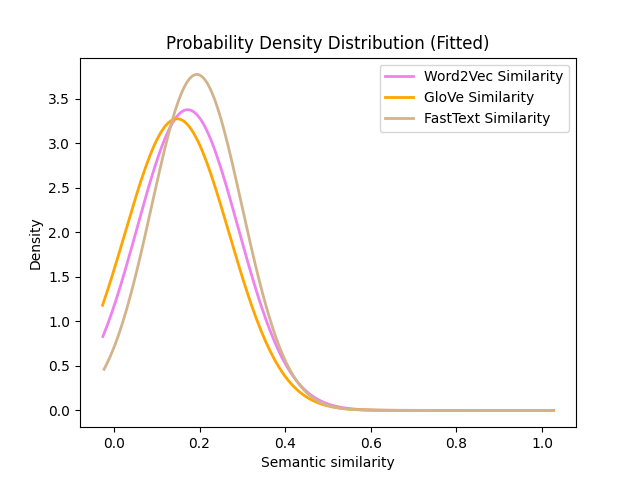}
			\end{minipage}
            \label{fig:pdf_te}
		}\hspace{-10pt}
		\hfill
		\subfloat[Pre-trained language model]{
			\begin{minipage}[b]{0.24\linewidth} 
				\centering
				\includegraphics[width=\linewidth,trim=20 20 25 25,clip]{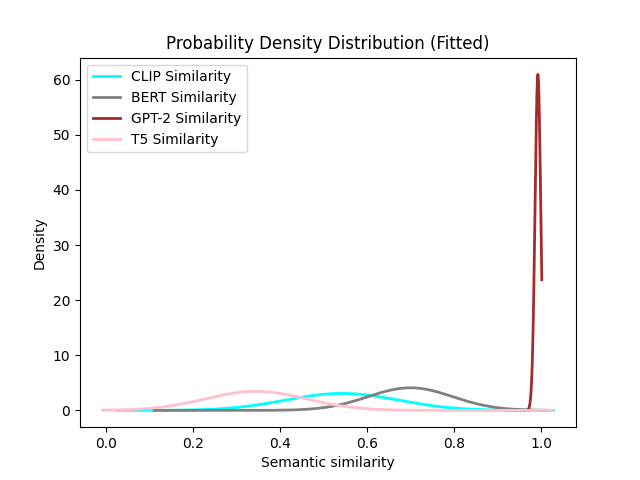}
			\end{minipage}
            \label{fig:pdf_lan}
		}
	\end{minipage}
    \vspace{-3mm}
	\caption{The fitted probability density distribution curves of similarity matrices for labels from the ImageNet-1k dataset, calculated using various similarity measurements, are presented as follows: Figure \ref{fig:pdf_all} displays all the methods under investigation, Figure \ref{fig:pdf_wn} showcases methods derived from WordNet, Figure \ref{fig:pdf_te} presents methods originating from text embedding models, and Figure (\ref{fig:pdf_lan}) illustrates methods derived from pre-trained language models.}
    \vspace{-2mm}
\label{fig:pdf}
\end{figure*}

\begin{figure*}[!t]
    \centering
    \includegraphics[width=0.92\textwidth, ,trim=8 2 5.5 2,clip]{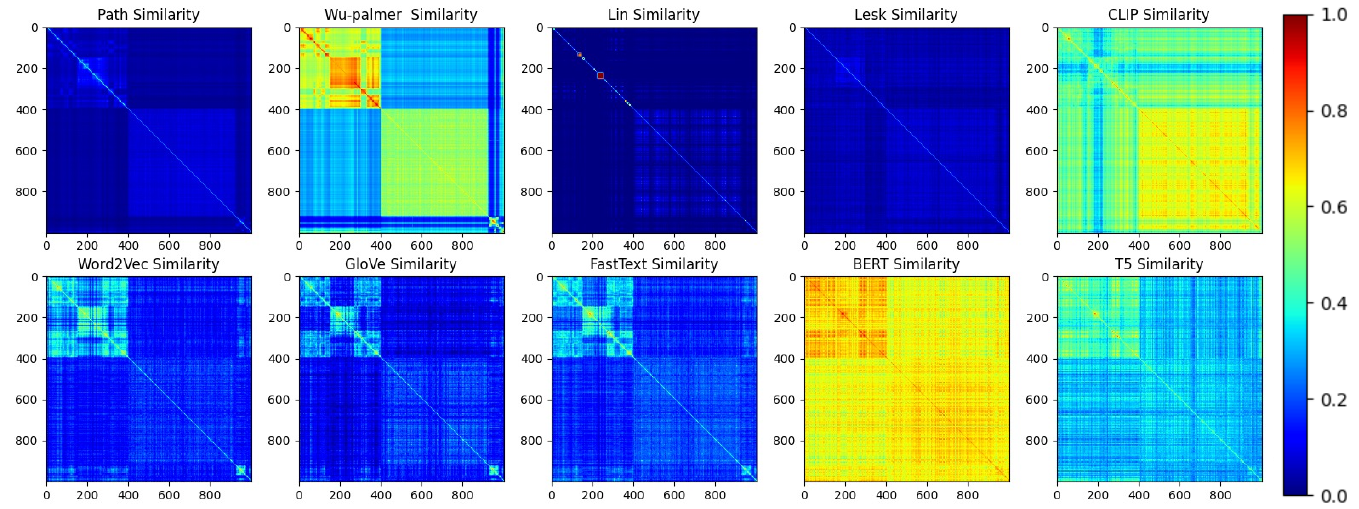}
    \vspace{-3mm}
    \caption{Heatmaps depicting similarity matrices for labels from the ImageNet-1k dataset, computed using various similarity measurements. Since a significant portion of values in the similarity matrix calculated by GPT-2 are close or equal to one, we do not present its heatmap here.}
    \label{fig:heatmap}
    \vspace{-2mm}
\end{figure*}

\begin{table}[!t]
    \centering
    \resizebox{\linewidth}{!}{
    \begin{tabular}{c|cccc|ccc|cccc}
        \toprule
         \multirow{2}*{Method} & \multicolumn{4}{c|}{WordNet method} & \multicolumn{3}{c|}{Text embedding model} & \multicolumn{4}{c}{Pre-trained language model} \\
         ~ & Lin & Lesk & Path & WuP & Glove & WV & FT & T5  & CLIP & BERT & GPT-2 \\
         \midrule
         Mean & 0.015 & 0.048 & 0.083 & 0.445 & 0.148 & 0.173 & 0.194 & 0.346 & 0.543 & 0.700 & 0.992 \\
         \midrule
         Std & 0.078 & 0.050 & 0.043 & 0.156 & 0.125 & 0.121 & 0.109 & 0.117 & 0.130 & 0.098 & 0.007 \\
         \bottomrule
    \end{tabular}}
    \vspace{-3mm}
    \caption{Mean and standard deviation values of the similarity matrices calculated for labels from the ImageNet-1k dataset using various similarity measurements. Herein, `WV,' `FT,' and `WuP' represent abbreviations for the Word2Vec, FastText, and Wu-Palmer methods.}
    \vspace{-2mm}
    \label{tab:mean_std}
\end{table}

\begin{figure}[!t]
    \centering
    \includegraphics[width=\linewidth, ,trim=20 15 10 10,clip]{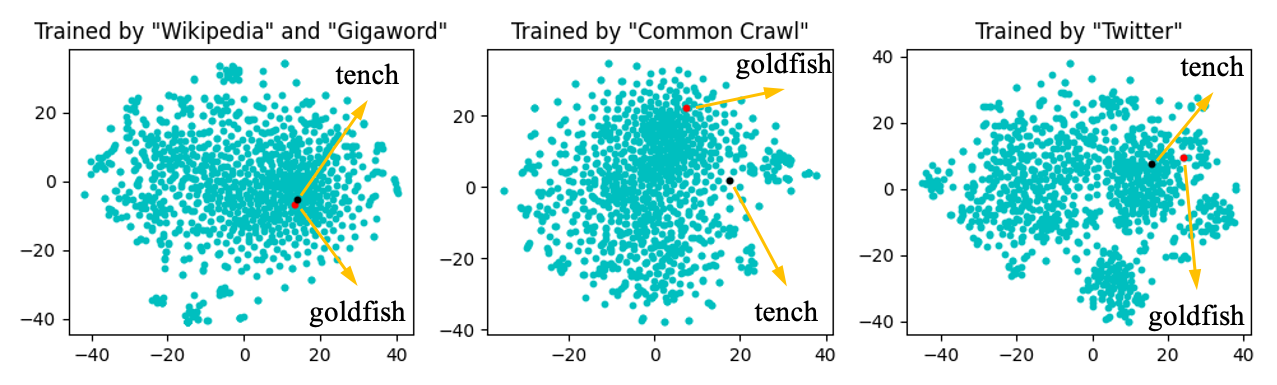}
    \vspace{-3mm}
    \caption{The t-SNE visualization of text embeddings for labels from the ImageNet-1k dataset, extracted from GloVe models trained on different datasets. The text embeddings that trained by different datasets have different distributions. The labels ``tench'' and ``goldfish'' display varying semantic distances within these distributions.}
    \label{fig:t_SNE}
    \vspace{-3mm}
\end{figure}

\begin{figure*}[!t]
    \centering
    \includegraphics[width=0.98\textwidth, ,trim=0 2 0 2,clip]{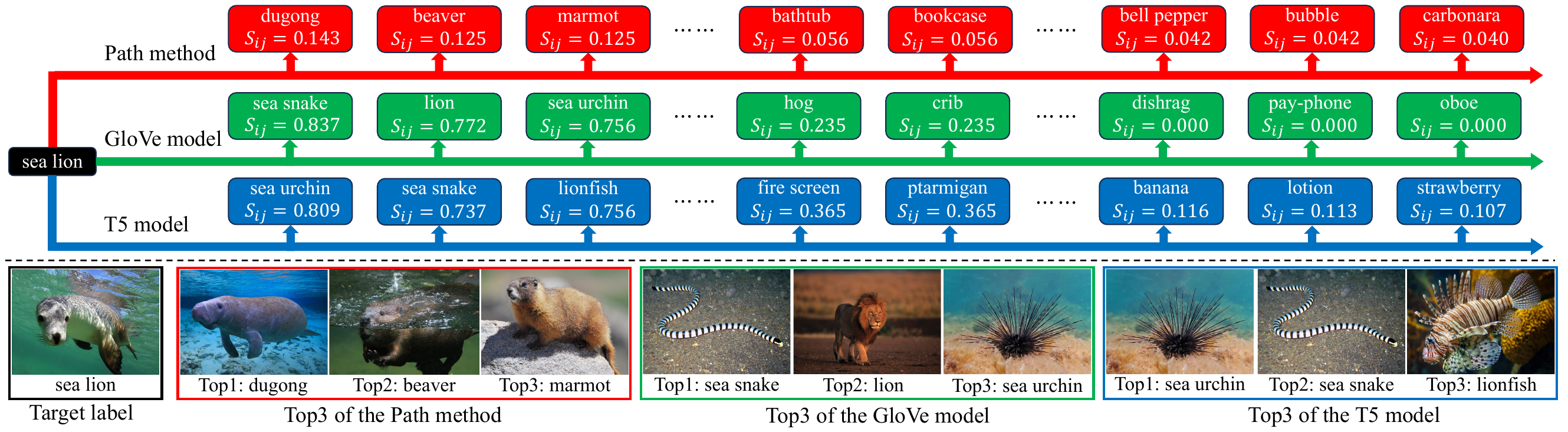}
    \vspace{-3mm}
    \caption{Similarity lists of one typical label ``sea lion'' calculated by the Path method, GloVe model, and T5 model. The GloVe and T5 models calculate large similarities to labels with similar text such as ``sea'' or ``lion''. While the Path method assigns large values to labels with inherent biological similarity. Besides, the GloVe and T5 models overestimate the semantic relationship between labels compared to the Path method of WordNet.}
    \label{fig:case_study}
    \vspace{-3mm}
\end{figure*}

\begin{figure}[t!]
	\centering
	\begin{minipage}[b]{0.95\linewidth}
		\subfloat[Crane: a bird]{
			\begin{minipage}[b]{0.46\linewidth} 
				\centering
				\includegraphics[width=\linewidth,trim=0 35 0 10,clip]{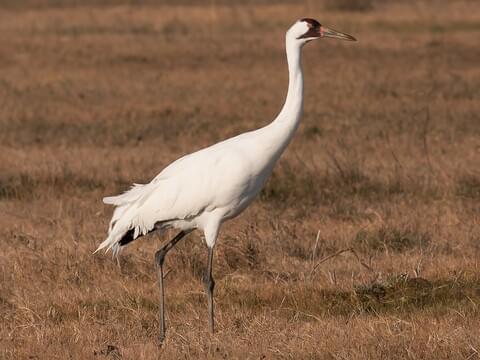}
			\end{minipage}
            \label{fig:crane_bird}
		}\hspace{-5pt}
		\hfill
		\subfloat[Crane: a machine]{
			\begin{minipage}[b]{0.46\linewidth}
				\centering
				\includegraphics[width=\linewidth,trim=0 12 0 0,clip]{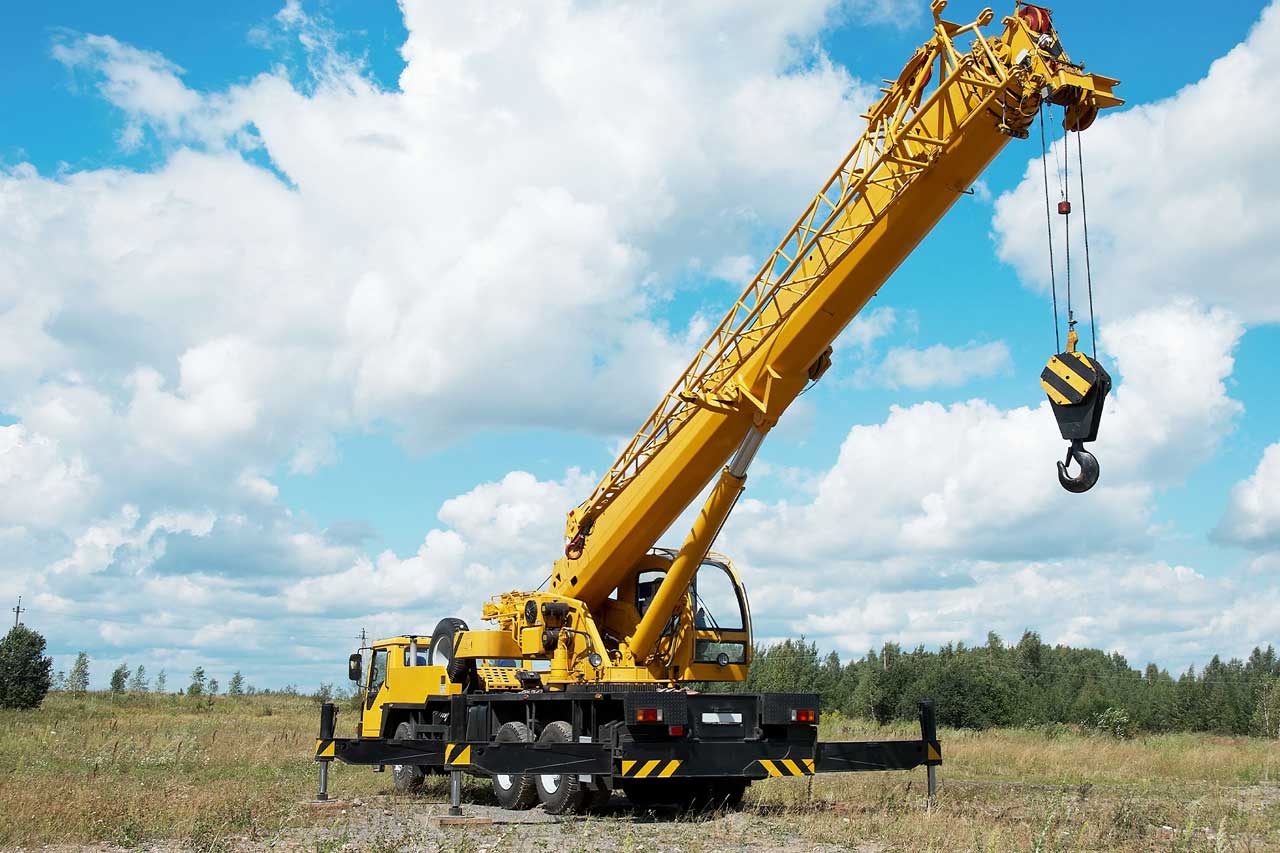}
			\end{minipage}
            \label{fig:crane_machine}
		}
	\end{minipage}
    \vspace{-3mm}
	\caption{The word ``crane'' with different meanings denotes the labels of a bird and a machine, respectively. Text embedding and pre-trained language models extract the same embedding for them and thus cannot distinguish them.}
    \vspace{-3mm}
\label{fig:crane}
\end{figure}

\begin{figure}[!t]
    \centering
    \includegraphics[width=\linewidth, ,trim=43.6 12 30 30,clip]{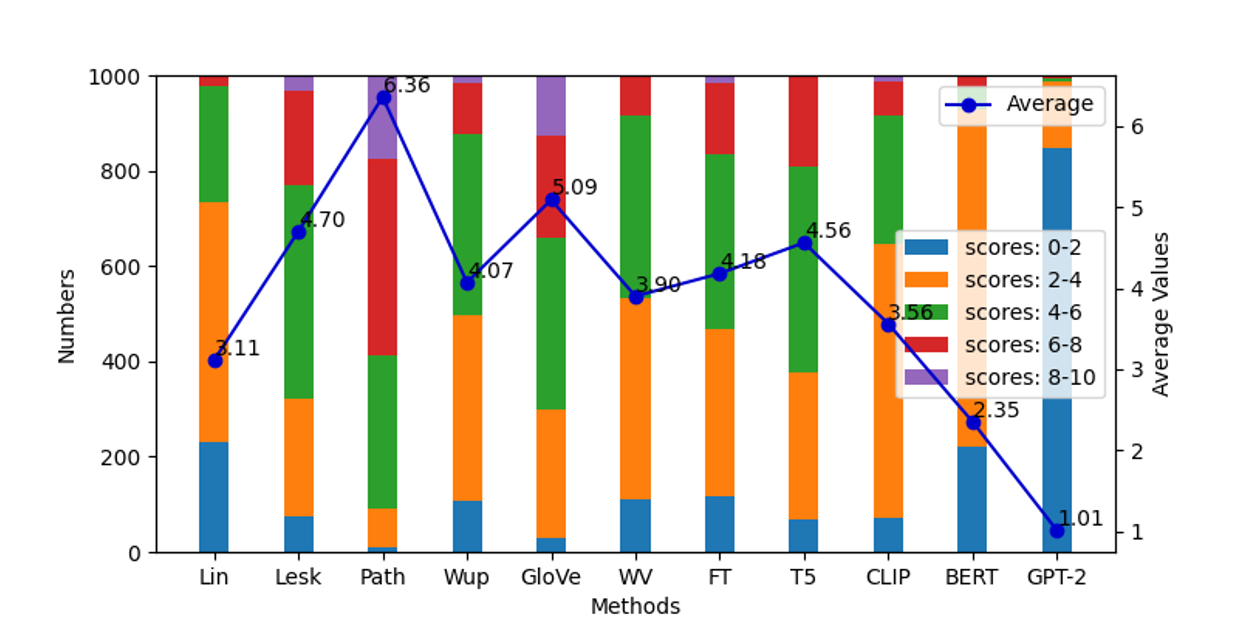}
    \vspace{-4mm}
    \caption{The visualization displays the distributions and mean scores resulting from a user study involving 1000 college students. This study encompassed 50 selected pairs of labels and their corresponding similarity values calculated using different methods. The scores range from 1 to 10, with 10 indicating the highest level of satisfaction.} 
    \label{fig:user_study}
    \vspace{-3mm}
\end{figure}

In this section, we analyze the eleven similarity measurements that belong to WordNet methods (Path \cite{lesk1986automatic}, Wu-Palmer \cite{wu1994verb}, Lin \cite{lin1998information}, and Lesk \cite{lesk1986automatic}), text embedding models (Word2Vec \cite{mikolov2013distributed}, GloVe \cite{pennington2014glove}, and FastText \cite{bojanowski2017enriching}), and pre-trained language models (CLIP \cite{radford2021learning}, BERT \cite{devlin2018bert}, GPT-2 \cite{radford2019language}, and T5 \cite{raffel2020exploring}). To analyze these methods, we use them to generate different similarity matrices for the labels in ImageNet-1k. We visualize the fitted probability density distribution curves and heatmaps of these similarity matrices in Figures \ref{fig:pdf} and \ref{fig:heatmap}, respectively. Furthermore, we report the mean and standard deviation values of these matrices in Table \ref{tab:mean_std}. Additionally, we conducted a user study to investigate human preferences, as visualized in Figure \ref{fig:user_study}.

\noindent
\textbf{WordNet Methods.} Table \ref{tab:mean_std} shows that Lin and Lesk methods have mean values of 0.015 and 0.048. So, as Figure \ref{fig:pdf_wn} shows, most similarity values of these two methods are distributed from 0 to 0.1, indicating that a significant portion of the similarity values is close or equal to zero. The heatmaps of Lin and Lesk, which are visualized in Figure \ref{fig:heatmap} validated this observation. hence, they underestimate the semantic relationship between words. Conversely, the Wu-Palmer method demonstrates a mean of 0.445, thereby overestimating the semantic relationship. The heatmap of Wu-Palmer, as visualized in Figure \ref{fig:heatmap}, shows that a substantial proportion of the similarity values exceed 0.5, further supporting this observation. In such cases, the proposed open metrics would yield evaluation results with significant bias compared to the vanilla metrics. 

\noindent
\textbf{Text Embedding Models.} Figures \ref{fig:pdf_te} and \ref{fig:heatmap}, and Table \ref{fig:heatmap} indicate that Word2Vec, GloVe, and FastText have a similar distribution, and the moderate mean values of them make them suitable choices. However, they have two major defects in calculating similarity. Firstly, text embedding models treat words as context-independent entities, which means they may not distinguish between different meanings of a word, leading to potentially incorrect similarity scores. For instance, in Figure \ref{fig:crane}, the word ``crane" in ImageNet-1k respectively represents the label of a bird and a machine, but text embedding models extract the same embedding for them. However, polysemy is a universal phenomenon in open-vocabulary tasks. In contrast, the base unit of WordNet is synset that organized by the meanings of words, which can inherently handle polysemy. Secondly, the quality of text embeddings heavily relies on the quality and size of the training corpus. Inadequate or biased training data can lead to biased embeddings. For example, Figure \ref{fig:t_SNE} illustrates that text embeddings from GloVe models trained on different datasets exhibit varying distributions. The labels ``tench" and ``goldfish" exhibit differing semantic distances within these distributions. However, WordNet is a structured network organized by thousands of linguists, psycholinguistics, and researchers in NLP over the past three decades \cite{fellbaum2010princeton}, which inherently mitigates biases in word representations.

\noindent
\textbf{Pre-trained Language Models.} We also use pre-trained language models to extract text embeddings. Therefore, the two major defects in text embedding models, \textit{i.e.} cannot handle polysemy and training data dependency, also exist in pre-trained language models. Besides, Figures \ref{fig:pdf_lan} and \ref{fig:heatmap}, and Table \ref{tab:mean_std} reveal that the similarity matrices calculated by pre-trained language models have large mean values, making them overestimate the semantic relationship. 

\noindent
\textbf{Case Study.} In Figure \ref{fig:case_study}, we visualize three similarity lists of one typical label ``sea lion'' in ImageNet-1k that calculated by the Path Similarity of WordNet, text embedding model GloVe, and T5 pre-trained language model. The similarity values of each word are sorted in descending order. For each method, we present the three largest, the two middle, and the three smallest values. We also attach the pictures of the labels of Top3 similarity values for each method. Figure \ref{fig:case_study} shows that the GloVe and T5 models calculate large similarities to labels containing similar text, such as ``sea'' or ``lion'' but with dissimilar appearances. In contrast, the Path method assigns large values to labels with inherent biological similarity and similar appearances. Besides, the GloVe and T5 models calculate similarity values far larger than the Path method, which obviously overestimates the semantic relationship between labels. 

\noindent
\textbf{Discussion.} The analyses above verify that Path Similarity has four advantages in calculating the similarity matrix for open-vocabulary tasks: 1) generates moderate similarity values. 2) able to distinguish polysemy. 3) unbiased to training data. 4) more aligned with human cognition. We also conducted a user study to assess human preferences, as depicted in Figure \ref{fig:user_study}. The results indicate that the Path method is significantly more favored than other methods. Consequently, based on our analysis, we prefer the Path method for calculating similarity. 

%% file: sec/5_benchmarks.tex
\section{Benchmarks}
\label{sec:benchmarks}

\begin{table}[!t]
	\centering
    \setlength{\tabcolsep}{2pt}
	\resizebox{\linewidth}{!}{
		\begin{tabular}{cc|ccccc}
			\toprule
			Method & Metric & A-847 & PC-459 & A-150 & PC-59 & PAS-20 \\
			\midrule
			\multirow{2}*{SimSeg \cite{xu2022simple}} & Vanilla mIoU & 6.7 & 8.7 & 20.4 & 47.3 & 88.3\\
			~ & Open mIoU & 12.9$^{\color{blue}(+6.2)}$ & 14.5$^{\color{blue}(+5.8)}$ & 29.0$^{\color{blue}(+8.6)}$ & 50.8$^{\color{blue}(+3.5)}$ & 90.2$^{\color{blue}(+1.9)}$\\
			\midrule
			\multirow{2}*{POMP \cite{ren2023prompt}} & Vanilla mIoU & - & - & 20.7 & 51.0 & -\\
			~ & Open mIoU & - & - & 29.6$^{\color{blue}(+8.9)}$ & 54.5$^{\color{blue}(+3.5)}$ & -\\
			\midrule
			\multirow{2}*{OVSeg \cite{liang2022open}} & Vanilla mIoU & 8.9 & 12.3 & 29.6 & 55.8 & 94.7\\
			~ & Open mIoU & 15.7$^{\color{blue}(+6.8)}$ & 17.4$^{\color{blue}(+5.1)}$ & 36.9$^{\color{blue}(+7.3)}$ & 59.6$^{\color{blue}(+3.8)}$ & 95.6$^{\color{blue}(+1.1)}$\\
			\midrule
			\multirow{2}*{SAN \cite{xu2023side}} & Vanilla mIoU & 12.8 & 16.6 & 31.9 & 57.5 & 95.2\\
			~ & Open mIoU & 19.2$^{\color{blue}(+6.4)}$ & 19.9$^{\color{blue}(+3.3)}$ & 39.0$^{\color{blue}(+7.1)}$ & 60.4$^{\color{blue}(+2.9)}$ & 96.0$^{\color{blue}(+0.8)}$\\
			\midrule
			\multirow{2}*{CAT-Seg \cite{cho2023cat}} & Vanilla mIoU & 11.4 & 16.2 & 31.5 & 61.6 & 96.5\\
			~ & Open mIoU & 18.4$^{\color{blue}(+7.0)}$ & 20.3$^{\color{blue}(+4.1)}$ & 39.9$^{\color{blue}(+8.4)}$ & 65.4$^{\color{blue}(+3.8)}$ & 97.2$^{\color{blue}(+0.7)}$\\
			\midrule
			\multirow{2}*{FC-CLIP \cite{yu2023convolutions}} & Vanilla mIoU & 14.8 & 13.8 & 34.1 & 58.4 & 95.4\\
			~ & Open mIoU & 20.6$^{\color{blue}(+5.8)}$ & 16.3$^{\color{blue}(+2.5)}$ & 40.4$^{\color{blue}(+6.3)}$ & 61.3$^{\color{blue}(+2.9)}$ & 96.2$^{\color{blue}(+0.8)}$\\
			\bottomrule
	\end{tabular}}
    \vspace{-3mm}
    \caption{Results on open-vocabulary semantic segmentation evaluated by vanilla and Open mIoU under a cross-dataset setting.}
	\vspace{-3mm}
\label{tab:ovss}
\end{table}

\begin{table}[!t]
	\centering
	\setlength{\tabcolsep}{2pt}
	\resizebox{\linewidth}{!}{
		\begin{tabular}{cc|ccc|ccc}
			\toprule
			\multirow{2}*{Method} & \multirow{2}*{Metric} & \multicolumn{3}{c|}{Source Dataset: COCO} & \multicolumn{3}{c}{Target Dataset: ADE20K (A-150)}\\
			~ & ~ & PQ & SQ & RQ & PQ &  SQ & RQ\\
			\midrule
			\multirow{2}*{OPSNet \cite{chen2023open}} & Vanilla PQ & 44.0 & 83.5 & 62.1 & 17.8 & 54.6 & 21.6\\
			~ & Open PQ & 45.9$^{\color{blue}(+1.9)}$ & 83.4$^{\color{red}(-0.1)}$ & 63.4$^{\color{blue}(+1.3)}$ & 21.9$^{\color{blue}(+4.1)}$ & 76.0$^{\color{blue}(+21.4)}$ & 26.7$^{\color{blue}(+6.1)}$\\
			\midrule
			\multirow{2}*{ODISE \cite{xu2023open}} & Vanilla PQ & 34.9 & 84.3 & 53.5 & 23.2 & 74.4 & 27.9\\
			~ & Open PQ & 37.1$^{\color{blue}(+2.2)}$ & 84.1$^{\color{red}(-0.2)}$ & 55.4$^{\color{blue}(+1.9)}$ & 25.6$^{\color{blue}(+2.4)}$ & 79.4$^{\color{blue}(+5.0)}$ & 30.9$^{\color{blue}(+3.0)}$\\
			\midrule
			\multirow{2}*{FC-CLIP \cite{yu2023convolutions}} & Vanilla PQ & 54.4 & 83.0 & 64.8 & 26.8 & 71.6 & 32.3\\
			~ & Open PQ & 55.3$^{\color{blue}(+0.9)}$ & 83.0 & 65.9$^{\color{blue}(+1.1)}$ & 28.8$^{\color{blue}(+2.0)}$ & 77.9$^{\color{blue}(+6.3)}$ & 34.8$^{\color{blue}(+2.5)}$\\
			\bottomrule
	\end{tabular}}
    \vspace{-3mm}
    \caption{Results on open-vocabulary panoptic segmentation evaluated by vanilla and Open PQ under a cross-dataset setting.}
	\vspace{-3mm}
\label{tab:ovps_crossdata}
\end{table}

\begin{table}[!t]
	\centering
	\setlength{\tabcolsep}{2pt}
	\resizebox{\linewidth}{!}{
		\begin{tabular}{cc|ccc|ccc}
			\toprule
			\multirow{2}*{Method} & \multirow{2}*{Metric} & \multicolumn{3}{c|}{Known} & \multicolumn{3}{c}{Unknow}\\
			~ & ~ & PQ$^{\rm Th}$ & SQ$^{\rm Th}$ & RQ$^{\rm Th}$ & PQ$^{\rm Th}$ &  SQ$^{\rm Th}$ & RQ$^{\rm Th}$\\
			\midrule
			\multirow{2}*{CGG \cite{wu2023betrayed}} & Vanilla PQ & 48.1 & 82.4 & 57.8 & 34.6 & 77.6 & 40.9\\
			~ & Open PQ & 48.8$^{\color{blue}(+0.7)}$ & 82.3$^{\color{red}(-0.1)}$ & 58.6$^{\color{blue}(+0.8)}$ & 35.3$^{\color{blue}(+0.7)}$ & 81.9$^{\color{blue}(+4.3)}$ & 41.8$^{\color{blue}(+0.9)}$\\
			\bottomrule
	\end{tabular}}
    \vspace{-3mm}
    \caption{Results on open-vocabulary panoptic segmentation under an in-dataset setting. Since only the "thing" classes are classified into "known" and "unknown" classes, we will only present the results for the "thing" classes here.}
	\vspace{-3mm}
\label{tab:ovps_indata}
\end{table}

\begin{table}[!t]
	\centering
	\setlength{\tabcolsep}{2pt}
	\resizebox{\linewidth}{!}{
		\begin{tabular}{cc|cccccc}
			\toprule
			\multirow{2}*{Method} & \multirow{2}*{Metric} & \multicolumn{6}{c}{Target Dataset: ADE20k (A-150)} \\
			~ & ~ & AP & AP50 & AP75 & AP$_s$ & AP$_m$ &  AP$_l$ \\
			\midrule
			\multirow{3}*{ODISE \cite{xu2023open}} & Vanilla AP & 14.0 & 23.4 & 13.9 & 3.3 & 16.2 & 27.6\\
			~ & Vanilla AP$^{\dag}$ & 12.7 & 20.9 & 12.7 & 2.8 & 14.5 & 25.3\\
			~ & Open AP$^{\dag}$ & 14.3$^{\color{blue}(+1.6)}$ & 23.7$^{\color{blue}(+2.8)}$ & 14.2$^{\color{blue}(+1.5)}$ & 3.8$^{\color{blue}(+1.0)}$ & 16.4$^{\color{blue}(+2.0)}$ & 27.3$^{\color{blue}(-2.0)}$\\
			\midrule
			\multirow{3}*{FC-CLIP \cite{yu2023convolutions}} & Vanilla AP & 16.8 & 27.4 & 17.2 & 5.7 & 19.9 & 29.7\\
            ~ & Vanilla AP$^{\dag}$ & 15.8 & 25.5 & 16.3 & 5.4 & 18.7 & 27.7\\
			~ & Open AP$^{\dag}$ & $17.1^{\color{blue}(+1.3)}$ & $27.8^{\color{blue}(+2.3}$ & $17.4^{\color{blue}(+1.1)}$ & $6.1^{\color{blue}(+0.7)}$ & $20.0^{\color{blue}(+1.3)}$ & $29.2^{\color{blue}(+1.5)}$\\
			\bottomrule
	\end{tabular}}
    \vspace{-3mm}
    \caption{Results on open-vocabulary instance segmentation evaluated by vanilla and Open AP under a cross-dataset setting. $*^{\dag}$ signifies the use of a class-agnostic global ranking during the matching process.}
	\vspace{-3mm}
\label{tab:ovis_crossdata}
\end{table}

\begin{table}[!t]
	\centering
	\setlength{\tabcolsep}{2pt}
	\resizebox{\linewidth}{!}{
		\begin{tabular}{cc|cc|ccc}
			\toprule
			\multirow{3}*{Method} & \multirow{2}*{Metric} & \multicolumn{2}{c|}{Constrained (AP50)} & \multicolumn{3}{c}{Generalized (AP50)}\\
			~ & ~ & Base & Novel & Base & Novel & All\\
			\midrule
			\multirow{3}*{CGG \cite{wu2023betrayed}} & Vanilla AP & 46.2 & 29.5 & 46.0 & 28.1 & 41.4\\
            ~ & Vanilla AP$^{\dag}$ & 44.2 & 28.9 & 43.6 & 26.4 & 39.1\\
			~ & Open AP$^{\dag}$ & 45.2$^{\color{blue}(+1.0)}$ & 30.0$^{\color{blue}(+1.1)}$ & 44.8$^{\color{blue}(+1.2)}$ & 28.7$^{\color{blue}(+2.3)}$ & 40.6$^{\color{blue}(+1.5)}$ \\
			\bottomrule
	\end{tabular}}
    \vspace{-3mm}
    \caption{Results on open-vocabulary instance segmentation under an in-dataset setting.}
	\vspace{-3mm}
\label{tab:ovis_indata}
\end{table}

\begin{table}[!t]
	\centering
	\setlength{\tabcolsep}{2pt}
	\resizebox{\linewidth}{!}{
		\begin{tabular}{cc|cc|ccc}
			\toprule
			\multirow{2}*{Method} & \multirow{2}*{Metric} & \multicolumn{2}{c|}{Constrained (AP50)} & \multicolumn{3}{c}{Generalized (AP50)}\\
			~ & ~ & Base & Novel & Base & Novel & All\\
			\midrule
			\multirow{3}*{CGG \cite{wu2023betrayed}} & Vanilla AP & 46.0 & 30.4 & 45.9 & 28.9 & 41.4\\
            ~ & Vanilla AP$^{\dag}$ & 44.3 & 29.4 & 43.6 & 26.9 & 39.2\\
			~ & Open AP$^{\dag}$ & 45.3$^{\color{blue}(+1.0)}$ & 30.6$^{\color{blue}(+1.2)}$ & 44.9$^{\color{blue}(+1.3)}$ & 29.1$^{\color{blue}(+2.2)}$ & 40.8$^{\color{blue}(+1.6)}$\\
			\midrule
			\multirow{3}*{RegionCLIP \cite{zhong2022regionclip}} & Vanilla AP & 62.1 & 42.9 & 61.7 & 39.3 & 55.9\\
			~ & Vanilla AP$^{\dag}$ & 61.0 & 39.8 & 60.3 & 30.1 & 52.4\\
			~ & Open AP$^{\dag}$ & 61.6$^{\color{blue}(+0.6)}$ & 40.6$^{\color{blue}(+0.8)}$ & 61.6$^{\color{blue}(+1.3)}$ & 33.6$^{\color{blue}(+3.5)}$ & 53.9$^{\color{blue}(+1.5)}$\\
			\midrule
			\multirow{3}*{Rasheed \textit{et al.} \cite{bangalath2022bridging}} & Vanilla AP & - & - & 56.7 & 40.5 & 52.5\\
            ~ & Vanilla AP$^{\dag}$ & - & - & 54.8 & 36.9 & 50.1\\
			~ & Open AP$^{\dag}$ & - & - & 55.8$^{\color{blue}(+1.0)}$ & 38.7$^{\color{blue}(+1.8)}$ & 51.3$^{\color{blue}(+1.2)}$\\
			\midrule
			\multirow{2}*{CORA \cite{wu2023cora}} & Vanilla AP & - & - & 44.7 & 41.5 & 43.9\\
            ~ & Vanilla AP$^{\dag}$ & - & - & 42.2 & 37.9 & 41.1\\
			~ & Open AP$^{\dag}$ & - & - & 43.6$^{\color{blue}(+1.4)}$ & 39.8$^{\color{blue}(+1.9)}$ & 42.6$^{\color{blue}(+1.5)}$\\
			\bottomrule
	\end{tabular}}
    \vspace{-3mm}
    \caption{Results on open-vocabulary object detection under an in-dataset setting.}
	\vspace{-3mm}
\label{tab:ovob_indata}
\end{table}

\begin{table}[!t]
	\centering
	\setlength{\tabcolsep}{2pt}
	\resizebox{\linewidth}{!}{
		\begin{tabular}{ccc|ccc}
			\toprule
			\multirow{2}*{Method} & \multirow{2}*{Metric} & \multirow{2}*{\makecell{Similarity \\ Measures}} & \multicolumn{3}{c}{Target Dataset: ADE20K (A-150)}\\
			~ & ~ & & PQ &  SQ & RQ\\
			\midrule
			\multirow{4}*{FC-CLIP \cite{yu2023convolutions}} & Vanilla PQ & - & 26.8 & 71.6 & 32.3\\
            \cline{2-6}
			~ & \multirow{3}*{Open PQ} & Path Similarity & 28.8$^{\color{blue}(+2.0)}$ & 77.9$^{\color{blue}(+6.3)}$ & 34.8$^{\color{blue}(+2.5)}$\\
			~ & ~ & GloVe model & 31.5$^{\color{blue}(+4.7)}$ & 77.7$^{\color{blue}(+6.1)}$ & 38.2$^{\color{blue}(+5.9)}$\\
			~ & ~ & T5 model & 30.7$^{\color{blue}(+3.9)}$ & 77.8$^{\color{blue}(+6.2)}$ & 37.1$^{\color{blue}(+4.8)}$\\
			\bottomrule
	\end{tabular}}
    \vspace{-3mm}
    \caption{Results on open-vocabulary panoptic segmentation were evaluated using Open PQ with three different similarity matrices calculated by the Path Similarity, GloVe model, and T5 model.}
	\vspace{-3mm}
\label{tab:ovps_ablation}
\end{table}

In this section, we employ Open mIoU, Open PQ, and Open AP metrics to assess open-vocabulary semantic (Sec \ref{sec.5.1}), panoptic (Sec \ref{sec.5.2}), and instance segmentation (Sec \ref{sec.5.3}), respectively. All results presented in this section are tested using officially released models from various methods. 

\subsection{Open-Vocabulary Semantic Segmentation}
\label{sec.5.1}

Table \ref{tab:ovss} provides the results on open-vocabulary semantic segmentation methods evaluated by vanilla and Open mIoU. All these methods are trained on the COCO-Stuff \cite{caesar2018coco} dataset and evaluated on challenging datasets such as Pascal VOC (PAS-20) \cite{mottaghi2014role}, Pascal Context (PC-59/459 for 59/459 categories) \cite{everingham2010pascal}, and ADE20K (A-150/847 for 150/846 categories) \cite{zhou2019semantic}. The Open mIoU consistently outperforms the vanilla mIoU across all five test datasets. This result aligns with expectations, as the Open mIoU assigns reasonable scores to error label predictions based on the similarity between predicted and ground truth labels. Additionally, as the vanilla mIoU scores increase, the improvement achieved by the Open mIoU becomes less pronounced. This discrepancy can be attributed to the lower occurrence of misclassified labels in predictions with higher vanilla mIoU scores. 


\subsection{Open-Vocabulary Panoptic Segmentation}
\label{sec.5.2}

Tables \ref{tab:ovps_crossdata} and \ref{tab:ovps_indata} present the results of open-vocabulary panoptic segmentation methods in both cross-dataset and in-dataset settings. In Table \ref{tab:ovps_crossdata}, all methods are trained on the source dataset COCO \cite{lin2014microsoft} and zero-shot evaluated on the target dataset ADE20K. Conversely, method in Table \ref{tab:ovps_indata} is trained and tested on the COCO dataset, where 64 thing classes are categorized as known classes, while the remaining 16 thing classes are designated as unknown classes.

\noindent
\textbf{Segmentation Quality.} The SQ in vanilla PQ only considers matched mask pairs, which may not reflect the segmentation quality accurately as it incorporates classification accuracy. In contrast, the SQ term in Open PQ evaluates the segmentation quality of overlapping masks ($\delta > 0.5$) regardless of their classification accuracy. In Table \ref{tab:ovps_crossdata}, Open SQ exhibits similar results as compared to vanilla SQ on the source dataset, suggesting that mask pairs with incorrect labels have comparable segmentation quality to those with correct labels. Conversely, on the target dataset, the Open SQ substantially increases to the vanilla SQ, indicating that mask pairs with incorrect labels exhibit significantly better segmentation quality than those with correct labels. The same phenomenon is also observable in Table \ref{tab:ovps_indata} when comparing known and unknown classes. Apparently, the vanilla SQ severely underestimates the methods' segmentation quality on the target/unknown dataset. Consequently, the SQ term in Open PQ proves sensitive to segmentation quality and can captures the segmentation quality of mask pairs with incorrect label predictions. 

\noindent
\textbf{Recognition Quality.} Table \ref{tab:ovps_crossdata} shows a disparity in the improvement of Open RQ between the source and target datasets. This discrepancy arises because the model trained on the source dataset tends to classify masks into the most accurate category rather than semantic similar categories when inferring on the corresponding source dataset. However, the target dataset contains numerous unseen classes, and in such cases, the model classifies these masks into semantically similar categories. Consequently, more substantial improvements are observed for Open RQ on the target dataset. This observation validates the sensitivity of the proposed Open RQ to the semantic distances between labels, enabling the evaluation of recognition quality that considers semantic similarity. 


\subsection{Open-Vocabulary Instance Segmentation}
\label{sec.5.3}

In Tables \ref{tab:ovis_crossdata} and \ref{tab:ovis_indata}, we present the results of open-vocabulary instance segmentation in both cross-dataset and in-dataset settings. In the cross-dataset setting, the methods are trained on the COCO dataset and tested on the ADE20K dataset. In the in-dataset setting, the methods undergo both training and evaluation exclusively on the COCO dataset, where 48 classes are categorized as base classes and 17 classes are classified as novel classes. Table \ref{tab:ovis_crossdata} highlights the superiority of Open AP over vanilla AP in all aspects, suggesting that the proposed Open AP is robust to variations in mask size and matching thresholds when assessing the semantic relationship between different object classes. Table \ref{tab:ovis_indata} exhibits a similar phenomenon as observed in Table \ref{tab:ovps_crossdata}, where there is a discrepancy in the increase of Open AP between base and novel classes. This observation also underscores the sensitivity of Open AP to the semantic relatedness between different labels. Consequently, Open AP proves to be a viable metric for evaluating open-vocabulary instance segmentation tasks.

\noindent
\textbf{Open-Vocabulary Object Detection.} 
The Open AP metric can also evaluate object detection by replacing the mask IoU with box IoU. Table \ref{tab:ovob_indata} presents the results for open-vocabulary object detection in a similar in-dataset setting as used in Table \ref{tab:ovis_indata}, revealing analogous observations to those in Table \ref{tab:ovis_indata}. This validates that the proposed Open AP is a reasonable metric for open-vocabulary object detection. Furthermore, the stability of Open AP improvements for CGG \cite{wu2023betrayed} in Tables \ref{tab:ovis_indata} and \ref{tab:ovob_indata} suggests that Open AP can be applied across different IoU types.

\begin{table}[!t]
    \centering
    \resizebox{0.75\linewidth}{!}{
    \begin{tabular}{c|ccc}
        \toprule
         Similarity methods & Path method & GloVe model & T5 model\\
         \midrule
         Mean & 6.12 & 3.05 &  4.29 \\
         \midrule
         Std & 1.71 & 0.99 & 1.23  \\
         \bottomrule
    \end{tabular}}
    \vspace{-3mm}
    \caption{The mean and standard deviation of the satisfaction scores obtained from a user study involving 1000 college students with 50 evaluated images are presented. The scores range from 1 to 10, with 10 indicating the highest level of satisfaction.} 
    \vspace{-2mm}
    \label{tab:use_study2}
\end{table}

\subsection{Ablation Study}
Table \ref{tab:ovps_ablation} presents the results for open-vocabulary panoptic segmentation evaluated by Open PQ with three different similarity matrices calculated by the Path Similarity, GloVe text embedding model, and language model T5. The increments in Open SQ exhibit similar values when different similarity matrices are used. This is expected because, in our designs, SQ only considers segmentation quality rather than classification accuracy, which is more reasonable. When using the GloVe or T5 model, the increases in Open RQ exhibit significantly larger values than the Path method, indicating an overestimated evaluation of classification accuracy for masks. The evaluation of open metrics yields moderate results when using the Path method, making it the most preferred choice for open-vocabulary tasks. Furthermore, we conducted a user study to assess human preferences when evaluating results using these different similarity matrices, as reported in Table \ref{tab:use_study2}. The findings indicate that humans are more inclined to accept evaluation results when assessed using the Path method.

%% file: sec/6_conclusion.tex
\section{Conclusion}
\label{sec:conclusion}

Our work addresses the limitations of existing evaluation methods in open-vocabulary image segmentation by proposing new open evaluation metrics that incorporate semantic similarity between predicted and ground truth labels. We emphasize the superiority of the Path Similarity method in calculating semantic similarity for open-vocabulary tasks, as it captures meaningful relationships between category labels. Through extensive evaluations, we demonstrate our proposed metrics' effectiveness in assessing models' performance on open-vocabulary image segmentation tasks. Our contributions enhance the evaluation framework for open-vocabulary segmentation, enabling better model development and assessment in real-world applications.